\documentclass[10pt,journal]{IEEEtranTCOM}

\usepackage[english]{babel}
\usepackage[usenames]{color}
\usepackage[cp1250]{inputenc}
\usepackage{amsfonts}
\usepackage{amsthm}
\usepackage{graphicx}
\usepackage{epsfig}
\usepackage{mathrsfs}
\usepackage{amsmath}
\usepackage{algorithm}
\usepackage{algorithmic}
\usepackage{amssymb}

\theoremstyle{plain}

\begin{document}
\newcommand{\bea}{\begin{eqnarray}}
\newcommand{\eea}{\end{eqnarray}}
\newcommand{\be}{\begin{equation}}
\newcommand{\ee}{\end{equation}}
\newcommand{\beas}{\begin{eqnarray*}}
\newcommand{\eeas}{\end{eqnarray*}}
\newcommand{\bs}{\backslash}
\newcommand{\bc}{\begin{center}}
\newcommand{\ec}{\end{center}}
\def\SC {\mathscr{C}}

\title{Hierarchical correlation reconstruction\\with missing data,\\for example for biology-inspired neuron}
\author{\IEEEauthorblockN{Jarek Duda}\\
\IEEEauthorblockA{Jagiellonian University,
Golebia 24, 31-007 Krakow, Poland,
Email: \emph{dudajar@gmail.com}}}
\maketitle

\begin{abstract}
Machine learning often needs to model density from a multidimensional data sample, including correlations between coordinates. Additionally, we often have missing data case: that data points can miss values for some of coordinates. This article adapts rapid parametric density estimation approach for this purpose: modelling density as a linear combination of orthonormal functions, for which $L^2$ optimization says that (independently) estimated coefficient for a given function is just average over the sample of value of this function. Hierarchical correlation reconstruction first models probability density for each separate coordinate using all its appearances in data sample, then adds corrections from independently modelled pairwise correlations using all samples having both coordinates, and so on independently adding correlations for growing numbers of variables using often decreasing evidence in data sample. A basic application of such modelled multidimensional density can be imputation of missing coordinates: by inserting known coordinates to the density, and taking expected values for the missing coordinates, or even their entire joint probability distribution. Presented method can be compared with cascade correlations approach, offering several advantages in flexibility and accuracy. It can be also used as artificial neuron: maximizing prediction capabilities for only local behavior - modelling and predicting local connections.
\end{abstract}
\section{Introduction}
Real world data points often miss values of some coordinates due to various reasons~\cite{missing}. This is a crucial issue for example for machine learning algorithms - often requiring numerical value in every position. Imputation tries to guess the missing values e.g. by averaging all occurrences of a given coordinate in data sample. However, it would neglect correlations with known coordinates of this data point. Additionally, imputation might have ambiguity requiring special attention, e.g. if sample comes from a circle and we know only one coordinate let say going through the center of this circle, imputation as expected value would be improper - we need entire density for proper conclusion, as discussed in Fig. \ref{circle}.

\begin{figure}[t!]
    \centering
        \includegraphics{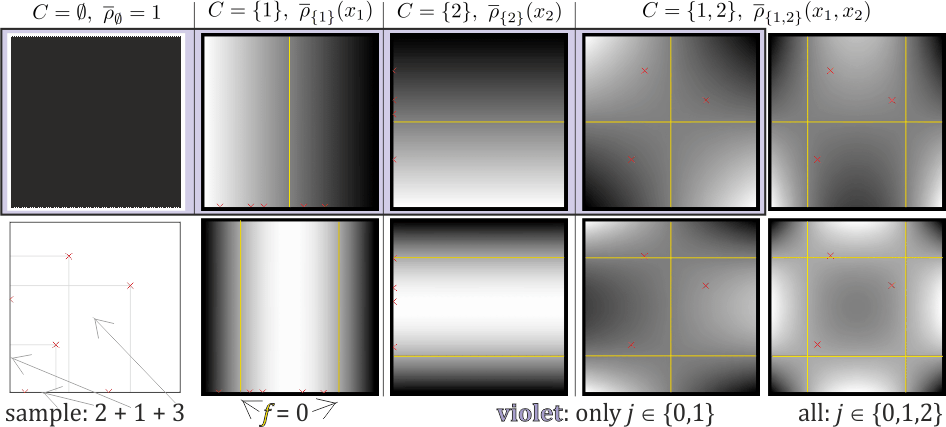}
        \caption{Orthonormal basis of polynomials for hierarchical correlation reconstruction in 2D ($d=2$): of density in $[0,1]^2$ as a linear combination of presented functions: $\rho(\textbf{x})=\sum_{C\subset \{1,2\}} \overline{\rho}_C(\textbf{x})=\sum_{f\in B} a_f f(\textbf{x})$. Data sample contains 3 points with known both coordinates, 2 points with known only the first coordinate and 1 point with known only the second coordinate - the remaining information is missing. In the upper-left corner is $\bar{\rho}_\emptyset=\rho_\emptyset=1$ initial density ensuring normalization to 1. The remaining graphs present basis  $f \in B=\bigcup_C B_C$ of polynomials we use: $f(\textbf{x})=f_{j_1}(x_1)\cdot f_{j_2}(x_2)$ for $j_i\in \{0,m\}$ (marked violet $m=1$, all for $m=2$). All but $\rho_\emptyset$ integrate to 0 (do not change normalization). Yellow lines show $f=0$. Estimation of $C=\{1\}$ coefficients requires to know only the first coordinate: we can use $2+3$ values. Analogously for $C=\{2\}$ we have $1+3$ values - still reconstructing densities of independent variables. Finally, $C=\{1,2\}$ basis allows to add corrections from pairwise correlations, requiring to know both coordinates. For example $f_{11}$ models increase/decrease of one variable while increasing second, $f_{12}$ models focus/spread of one variable while increasing second, and so on. Generally, $C$ correlations allow to imply behavior of any of $C$ coordinates from the remaining $|C|-1$.}
       \label{missingen}
\end{figure}

There will be proposed and discussed inexpensive general systematic way to reconstruct entire $d$-dimensional joint probability density function - including correlations, for example for more accurate data imputation. For simplicity, flexibility and the best use of all acquired data: $\{\textbf{x}^k\}_{k=1..n}$, $\textbf{x}^k\in\mathbb{R}^d$ with some coordinates missing, the process is made in hierarchical way as visualized in Fig. \ref{missingen}: start with reconstructing density function for each separate coordinate, what allows to use all appearances of given coordinate in data sample. For such initial density as product of independent random variables, we add corrections from correlations between all pairs of coordinates - using all data points having both coordinates. And so on - we can accumulate corrections from correlations of growing number of coordinates $C\subset \{1,\ldots,d\}$, using all data points having at least $C$ coordinates. Evidence for correlations between larger number of coordinates might decrease in incomplete data sample, so the discussed process allows to customize such modelling accordingly to given sample.
\begin{figure}[t!]
    \centering
        \includegraphics{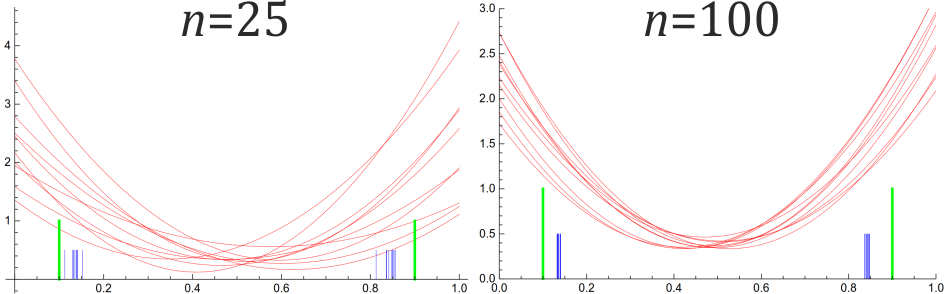}
        \caption{2D imputation example: for sample from uniform distribution in circle centered in $(0.5,0.5)$ and radius $0.4$. We would like to perform imputation of $x_1$ coordinate knowing that $x_2=0.5$. Expected value is $x_1=0.5$, but using it for imputation would be improper as the sample avoids this center of circle. Presented experiments fit $(L^2)$ density on $[0,1]^2$ as degree 2 polynomial (8 coefficients with basis as in Fig. \ref{missingen}) to random sample of size $n=25$ (left) or $n=100$ (right). Plots show $\rho((x_1,0.5))$ cross-sections: 10 red lines represent independent 10 experiments. After normalization, such cross-section can be used as estimated density for unknown coefficient $x_1$. As it is parabola, we can use its minimum to split into two clusters and take expected value for each cluster - they are drawn as the blue lines, while green line would be the perfect prediction.}
       \label{circle}
\end{figure}

Specifically, it adapts rapid parametric density estimation~\cite{rapid} approach for this missing data case (simple simulator:~\cite{sim}), which models density as a linear combination of chosen functions. Mean-square optimization allows for inexpensive calculation of estimated coefficients for such functions, especially if using orthonormal basis, for which coefficient for function $f$ can be independently estimated as just average of this function over the data sample: $\rho(\textbf{x})\approx \sum_j [f_j] f_j(\textbf{x})$, where $[f]=\frac{1}{n}\sum_k f(\textbf{x}^k)$.

A convenient orthonormal basis for multidimensional $[0,1]^d$ case are just products of functions from 1D orthonormal basis in $[0,1]$. Especially for the missing data case, this basis can be decomposed into functions representing only correlations for some subset of coordinates $C$, what allows to use all data points having known all $C$ coordinates as evidence to model this correlation - using a varying number of samples to estimate coefficient for $f$ as average over data sample.

Such simple automated system for modelling correlations between inputs, predicting given connection basing on the remaining ones, is analogous to behavior imagined for biological neurons e.g. in hierarchical temporal mmory~\cite{hawkins} - which are much more sophisticated than those used in standard artificial neural network. Having such promising analogue we could try to recreate neural networks with similar learning and reasoning mechanisms as biological counterparts. The flexibility of choosing basis $B=\bigcup_C B_C$, independently for correlations between variables $C$, allows to work both in dense regime: high order ($j=1\ldots m$) modeling of a few signals, or in sparse regime: low order modelling of long correlation patterns: $B_C$ functions allow to imply behavior of any of $C$ coordinates from the remaining $|C|-1$ variables. The discussed technique can be also directly compared with classical cascade correlation approach, offering much more flexibility, adaptivity and control of obtained results.
\section{Rapid parametric density estimation}
\begin{figure}[b!]
    \centering
        \includegraphics{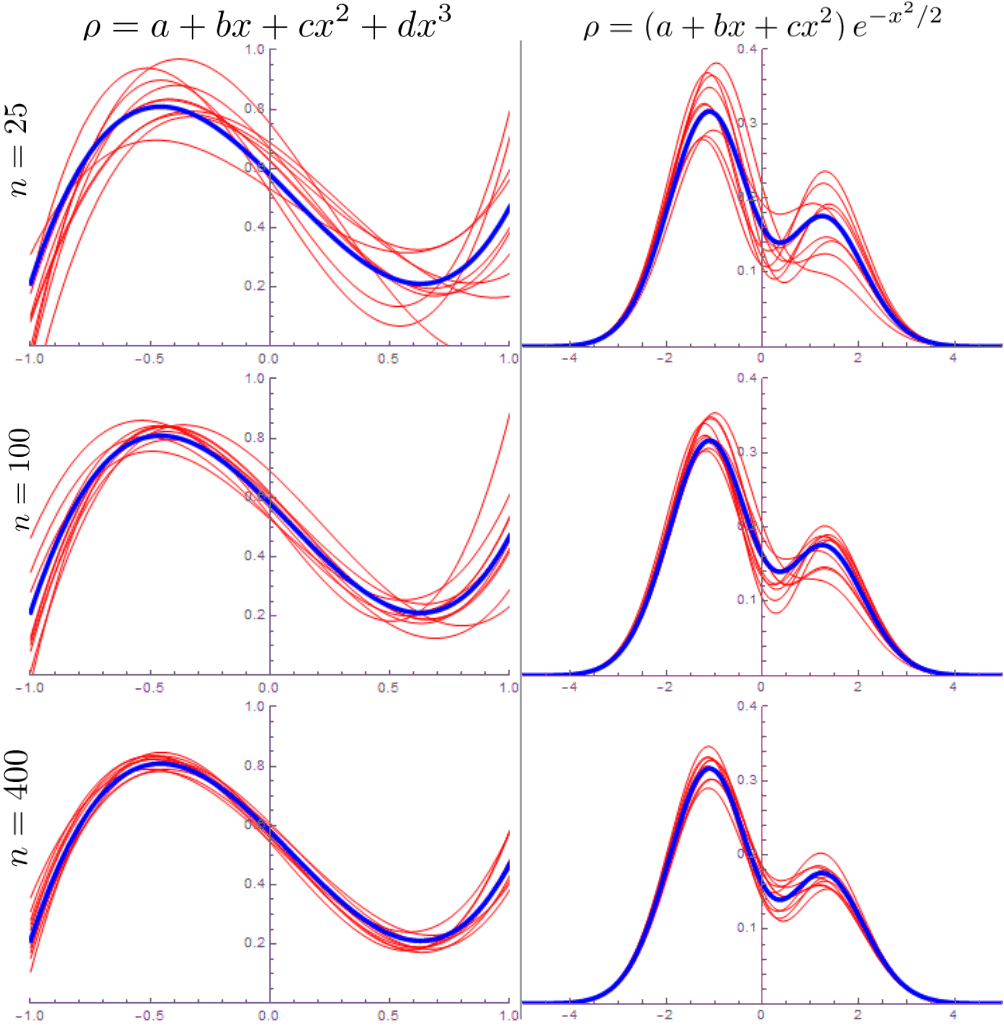}
        \caption{Two basic examples of 1D rapid parametric density estimation~\cite{rapid}: fitting polynomial (left column) in $[-1,1]$ range and polynomial multiplied by $e^{-x^2/2}$ (right) in $\mathbb{R}$, basing on a random sample of size $n=25$ (top row), $n=100$ (middle) or $n=400$ values (bottom), generated using the assumed probability distribution - represented as the thick blue line. Every plot contains also 10 thin red lines representing results of 10 independent experiments of estimating the parameters basing on the obtained size $n$ sample. Inaccuracy drops with $1/\sqrt{n}$, what can be seen in dispersion dropping approximately twice every row. For convenience there will be discussed using orthogonal family of functions, for example polynomials, making their parameters independent - calculated as just average of the value of a given function over the obtained sample. The polynomial formula used for the left column can be also expressed as $\rho=\frac{5}{8}\left(1-3x^2+[x](15x-21x^3)+[x^2](9x^2-3)+[x^3](35x^3-21x)\right)$,
        where $[f]$ denotes average of function $f$ over the sample.}
       \label{overv}
\end{figure}
Imagine we have a data sample $\{\textbf{x}^k\}_{k=1..n}$ and want to fit to it density as a linear combination of functions from some chosen basis $\rho=\sum_j a_j f_j$. A natural approach is smoothing the sample by convolution with width $\epsilon$ Gaussian kernel and optimizing parameters $a_j$ to minimize $L^2$ norm between such smoothed sample and estimated density $\rho$.

As discussed in \cite{rapid}, surprisingly, we get the best estimation of parameters if using $\epsilon\to 0$ limit (even though $L^2$ norm grows to infinity there). If we assume that the used family of functions is orthonormal: $\langle f_i, f_j\rangle =\int f_i(\textbf{x})f_j(\textbf{x}) d\textbf{x} = \delta_{ij}$, it turns out that optimal coefficient for $f_j$ is just average of $f_j$ over the sample:
\be \rho\approx \sum_j a_j f_j=\sum_j [f_j] f_j\quad\textrm{where}\quad [f]=\frac{1}{n}\sum_{k=1}^n f(\textbf{x}^k) \ee
Examples of such estimation using basis of polynomials (left) or polynomials multiplied by $e^{-x^2}$ (right) are shown in fig. \ref{overv}. It also visualizes that error of coefficients drops approximately like $1/\sqrt{n}$, what is standard behavior of uncertainty when estimating average. From the central limit theorem, the error of $j$-th coefficient comes from approximately normal distribution of width being standard deviation of $f_j$ (assuming it is finite) divided by $\sqrt{n}$:
\be [f_j]-a_j \sim \mathcal{N}\left(0,\frac{1}{\sqrt{n}}\sqrt{\int (f_j-a_j)^2 \rho\, d\textbf{x}}\right) \label{error}\ee

In our case we would like to construct orthonormal basis for multidimensional case as products of functions from orthonormal basis in 1D case. For simplicity, let us work on $[0,1]$ range and use $f_0=1$ as the initial function of our basis, ensuring independence from given coordinate - this coordinate missing in data point does not prevent using it to estimate given coefficient. Additionally, all the remaining $f_j$ have to integrate to 0: $\int_0^1 f_j(x)dx=\langle f_j,f_0\rangle =0$, hence $\rho_\emptyset=f_0\cdot\ldots\cdot f_0=1$ guards normalization to 1, further functions cannot cripple. However, they generally can damage in a different way: sometimes lead to negative densities - in many applications like imputation it can be practically  ignored, otherwise it should be handled or avoided.

\begin{figure}[t!]
    \centering
        \includegraphics{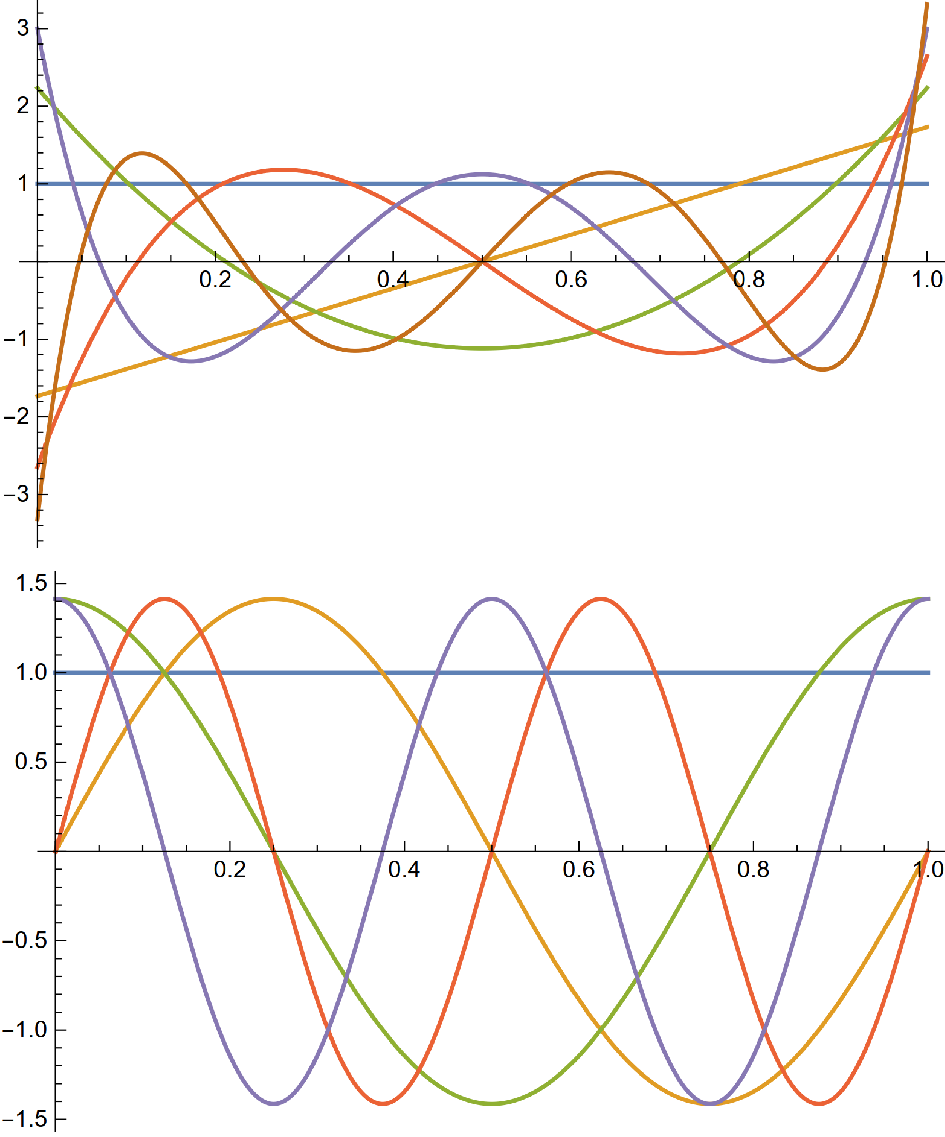}
        \caption{Plots of two basic choices for orthonormal basis for $\langle f, g\rangle =\int_0^1 f(x)g(x) dx$ scalar product. Both contain $f_0=1$, at the top there are polynomials (rescaled Legendre), at the bottom sines and cosines. Corrections using such basis have similar meaning as cumulants (correspondingly: expected value, variance, skewness, curtosis), however, are much more convenient for reconstruction of the probability distribution. }
       \label{zeroone}
\end{figure}

After necessary $f_0=1$, two basic ways to choose orthonomal family of functions for $[0,1]$ range are presented in Fig. \ref{zeroone}: (rescaled Legendre) polynomials - $f_1,\ldots,f_5$ correspondingly:
$$\sqrt{3}(2x-1), \sqrt{5}(6x^2-6x+1), \sqrt{7}(20x^3-30x^2+12x-1),$$
$$3(70x^4-140x^3+90x^2-20x+1), $$
$$\sqrt{11}(252x^5-630x^4+560x^3-210x^2+30x-1).$$

Alternative basic choice of orthonormal basis for $[0,1]$ are sines and cosines:
$$\sqrt{2} \sin(2\pi x \lceil j/2\rceil), \sqrt{2} \cos(2\pi x \lceil j/2\rceil).$$

There can be also considered different choices, including wavelets. From computational perspective, polynomials seem the most convenient, especially that cross-section after fixing some coordinates, derivatives, integrals are also polynomials.

For multidimensional case $[0,1]^d$ we can use products $f_{\textbf{j}}=f_{j_1}\cdot\ldots f_{j_d}$ e.g. for $\textbf{j}\in\{0,\ldots,m\}^d$ as orthonormal basis (generally different coordinates can use different functions), again independently estimating coefficients by averaging $f_{\textbf{j}}$ over given sample. For missing data case observe that $j_i=0$ coordinates do not depend on the values ($f_0=1$). Therefore, we can use data points missing these coordinates to estimate the coefficients - the averaging can be made over varying numbers of points: all having known at least $j_i>0$ positions. Basis to model correlations for $C$ coordinates ($f_\textbf{j}\in B_C$) have $j_i>0$ for $i\in C$ and $j_i=0$ for $i\notin C$.

\section{Hierarchical correlation reconstruction}
Assume we have a sample of $n$ points $\{\textbf{x}^k\}_{k=1..n}$ from $d$ dimensional space of numerical values, $D=\{1,\ldots,d\}$ is the set of coordinates. For convenience assume they are transformed to $x^k_i \in [0,1]$. It can be done by rescaling, or e.g. using $1/(1+e^{-x})$ for unbounded coordinates, or preferably with estimated cumulative distribution function for ensuring uniformity of separate variables. Some non-numerical coordinates can be interpreted as numerical, e.g. true/false as $1/0$, in which case basis functions restricted to these discrete values should form orthonormal vector basis.

In missing data case we allow some coordinates to be unknown. Denote by $C_k\subset D$ as known coordinates of $k$-th point: only for $i\in C_k$ there is known $x^k_i \in [0,1]$. The remaining coordinates are meaningless, cannot be used for density estimation.

For $C\subset D$ define $K_C=\{k:C\subset C_k\}$ as all points with known at least $C$ coordinates - being evidence for $C$ correlations. Denote $|K_C|$ as size of this subset, e.g. $|K_\emptyset|=n$, obviously: $C'\subset C \Rightarrow K_C \subset K_{C'}$.

Denote $\rho_C:[0,1]^d\to \mathbb{R}$ as density for coordinates $C$, ignoring the remaining coordinates (constant for them), only modeling density and correlations for coordinates $C$. We are mostly interested in the final: $\rho\equiv \rho_D$. It will be constructed in hierarchical way - by independently evaluating correlations in growing $C$: $\rho_C=\sum_{C'\subset C} \overline{\rho}_{C'}$, where $\overline{\rho}_{C'}$ is correlation contribution of $C'$ alone - not inflicting correlations for $C''\subsetneq C'$, like 4 functions for $\overline{\rho}_{\{1,2\}}$ in the right part of Fig. \ref{missingen}.

We start with $\overline{\rho}_\emptyset=\rho_\emptyset = \prod_i f_0 (x_i) =1$ uniform distribution on $[0,1]^d$. For $j\geq 1$ we have $\int_0^1 f_j(x)dx=\langle f_j,f_0\rangle =0$. Therefore, functions in $[0,1]^d$ orthonormal basis $f_{\textbf{j}}=f_{j_1}(x_1)\ldots f_{j_d}(x_d)$, having $j_i>0 \Leftrightarrow i\in C$ are perfect for basis of $\overline{\rho}_C$: taking marginal distribution (integrating) to a smaller $C'\subsetneq C$, we get zero.

Hence, for $\overline{\rho}_C$ contribution to $\rho=\sum_{C\subset D} \overline{\rho}_C$ we should choose basis:
\be B_C\subset \left\{f_{j_1}(x_1)\ldots f_{j_d}(x_d):\forall_{i\in C}\, j_i\geq 1,\forall_{i\notin C}\, j_i=0\right\} \ee
Choice of the basis $B=\bigcup_C B_C$ should be made based on data sample and available resources, can be also adapted on the way - increasing order of the model for a growing number of data samples, up to adding new coordinates, like while attaching new synapses to a neuron. It will be briefly discussed in the next section.

Having chosen orthogonal basis $B=\bigcup_{C\subset D} B_C $ (containing $\emptyset$) for $[0,1]^d$, coefficient for function $f\in B$ is just average of $f$ over the sample, however, in missing data case this averaging is over a varying amount of evidence: $|K_C|$ samples for correlation $C$:
\be \rho=\sum_{C\subset D} \overline{\rho}_C=\sum_{C\subset D}\sum_{f\in B_C} \frac{\sum_{k\in K_C} f(\textbf{x}^k)}{|K_C|} f \label{fin1}\ee
This sum starts with $\overline{\rho}_\emptyset=1$, then we model density for each separate coordinate $(|C|=1)$, then we independently model correlations between pairs of coordinates $(|C|=2)$, and so on - every time estimating coefficient from all points having at least $C$ coordinates. Without missing data we get $|K_C|=n$ standard case of averaging over all points, for which such hierarchical decomposition of correlations is also convenient.

\section{Remarks and modifications }
\subsection{Choosing the basis}
The discussed general approach should be customized for a given situation, starting with the choice of independently modelled functions for each coordinate: e.g. polynomials or sines/cosines, not necessarily the same for all coordinates. It is crucial to properly choose the basis $B=\bigcup_C B_C$: of modelled order of correlations for different sets of coordinates $C$. Some basic remarks:

\begin{itemize}
  \item The number of coefficients grows exponentially, e.g. $m^{|C|}$ if using all $j=1\ldots m$. It generally suggests reducing maximal order $m$ while modelling correlations for growing number of variables $|C|$, especially that it also reduces evidence in missing data case.
  \item For large $d$ we can use sparse basis: nonempty $B_C$ only for coordinates appearing together in our data sample, which turned out indeed correlated. A natural construction is starting with small $C$ sets and try to successively increase them. However, danger of such approach is that data looking like noise for a low order method, might have hidden higher order correlations. Therefore, a safer way is gathering information (averaging) for a large basis, but use only a smaller subset for the actual predictions: with more certain and significant coefficients. Larger values of basis functions for  separate coordinates $(f_{j_i}(x_i))$ might suggest more promising product functions $(f_{\textbf{j}}(\textbf{x}))$ to consider for sparse basis.
  \item From the other side, not using even $j=1$ in $B_C$ means that given behavior will be ignored, e.g. for $C=\{i\}$ it would means assuming uniform distribution for separate $i$-th coordinate, what is appropriate e.g. if using estimated cumulative distribution for transformation to $x_i\in[0,1]$. If some coordinates will be always known (e.g. inputs), we can omit modelling correlations between them.
  \item Having low evidence $|K_C|$, there is large $\propto |K_C|^{-1/2}$ uncertainty of coefficients - it might be safer not to use uncertain functions in the estimated density, especially that they are more likely to lead to problematic negative densities.
  \item Obtained small coefficients, especially comparing to their uncertainty, can be discarded from calculated density to reduce computational requirements. If coefficient remains insignificant while improving accuracy, it might be worth to consider this correlation irrelevant, reducing cost of adaptation and memory - for example replacing it with some yet unused function: additional type of correlation to independently model.
  \item For some data types it makes no sense to use high degree polynomials, e.g. for true/false values interpreted as $1/0$, it only makes sense to use linear $f_1$. If given coordinate can have only 3 values, there should be used at most $\{f_1,f_2\}$, and so on: at most $\{f_1,\ldots,f_m\}$ for coordinate obtaining $m+1$ values - naturally adapting discussed approach to discrete variables. In this case, functions from our basis become discrete vectors - we should ensure that these vectors form orthonormal basis.
\end{itemize}

\subsection{Adaptive averaging}
The (\ref{fin1}) formula can be written as $\rho=\sum_{f\in B} a_f f$, where
\be a_f=[f]=\frac{\sum_{k\in K_C} f(\textbf{x}^k)}{|K_C|}\qquad\textrm{for}\quad f\in B_C \label{avg1}\ee
is just average over the sample. It is appropriate for building final static model based on known entire sample. For local sensitivity or online adaptation to processed data (e.g. neuron model, data compression), it can be replaced with adaptive averaging: for some small learning rate $\lambda$, update coefficients accordingly to current observation $\textbf{x}$:
\be a_f \to (1-\lambda) a_f +\lambda f(\textbf{x}) \label{adapt}\ee
or equivalently $a_f\ +=\ \lambda(f(\textbf{x})-a_f)$. If observation $\textbf{x}$ has only coordinates $C$, the update is only for $f\in B_{C'}$ for all $C'\subset C$. The learning rate $\lambda$ can be chosen as large at the beginning, then reduced while stabilization, e.g. $0.05\to 0.001$. If chosen as a power of two, multiplication becomes less expensive bitshift.
\subsection{Imputation}
In imputation we want to predict missing coordinates of data points. A natural approach here is inserting the known coordinates into the modelled density $\rho$, after normalization getting joint probability distribution for the remaining coordinates. This normalization requires integration over these missing coordinates, what for our basis is quite simple:
$$\int_0^1 f_{j_1}\cdot\ldots\cdot f_{j_d} dx_i=0 \textrm{ if }j_i > 0,\textrm{ else}\ f_{j_1}\cdot\ldots \cdot f_{j_d}$$
For generality, assume we know $C$ coordinates and are interested in joint distribution for $\bar{C}\subset D\setminus C$ subset of the remaining coordinates (e.g. $\bar{C}= D\setminus C$). Obtained density restricted to known $C$ coordinates of $\textbf{x}$ is
\be \rho_{C,\bar{C}}(\textbf{x})=\frac{\sum_{f\in B_{C'}:C'\subset C\cup \bar{C}}\ a_f f(\textbf{x})}{\sum_{f\in B_{C'}:C'\subset C}\ a_f f(\textbf{x})} \ee
where $C$ coordinates of $\textbf{x}$ are fixed, $\bar{C}$ coordinates vary - above density is for these $\bar{C}$ coordinates of $\textbf{x}$. Denominator comes from integration over all $\bar{C}$ coordinates (it contains also functions from nominator, allowing to simplify above formula).

For imputation we are usually interested in probability density for separate single variables: for $i\notin C$, $\rho_{C,i}(\bar{x}_i)\equiv \rho_{C,\{i\}}((x_1\ldots \bar{x}_i\ldots x_d))$:
\be \rho_{C,i}(\bar{x}_i)=\frac{\sum_{f\in B_{C'}:C'\subset C\cup \{i\}}\ a_f\, f((x_1\ldots \bar{x}_i\ldots x_d))}{\sum_{f\in B_{C'}:C'\subset C}\ a_f \,f(\textbf{x})} \ee
Which is e.g. polynomial of single variable, like in Fig. \ref{circle}, from which we can easily calculate expected value (and e.g. variance for uncertainty):
\be E_{C,i}=\int_0^1 \bar{x}_i\, \rho_{C,i}(\bar{x}_i) d\bar{x}_i\qquad E_i\equiv E_{C\setminus \{i\},i} \label{exp}\ee

However, generally imputation might not always be unique, like for sample from a circle discussed in Fig. \ref{circle}: centered in $(0.5,0.5)$, of radius $0.4$. Knowing only $x_1=0.5$, imputation as expected value: $x_2=0.5$ does not agree with data. Instead, proper imputation should give $x_2=0.5 \pm 0.4$, for example taking one of these options, or maybe both: it might be safer to split such data point into a few: using alternative imputation choices, maybe giving lower weights to such split points.

Density modelled in the discussed approach can suggest such ambiguity problem through large variance. In this case, we can e.g. take a (global?) maximum of density instead, or maybe a few local maxima - preferably of joint probability density of all missing coordinates.

Considering such possibilities, we should ask a natural question if imputation in given case should return expected value, or maybe rather e.g. global maximum of density? Such maximum can be narrow: contain low probability, so maybe we should cluster density instead and return center (expected value in Fig. \ref{circle}) of the most probable cluster? In higher dimensions we can split this question into successively choosing single coordinates using 1D probability densities: start with coordinate giving the best certainty and use the earlier choices during the following ones. Presented modelling of density as just polynomial allows to work with various choices for this question, or can even just return modelled joint probability distribution for all the missing coordinates.

\subsection{Improving likelihood}
Discussed mean-square optimization is often a natural choice, and allows for inexpensive and independent calculation of coefficients. However, for probabilistic interpretation, it is often preferred to use likelihood optimization instead.  Hence, especially for small data samples, it might be worth to complement previous estimation with a few steps of e.g. gradient descent, improving log-likelihood of initial $\rho=\sum_f a_f f$: modify parameters to increase probability of obtaining the given sample. It introduces dependencies between previously independent coefficients.

Normalization depends only on $\rho_\emptyset = 1$, which coefficient has to be 1. The remaining can be freely modified, however, it might lead to negative densities. Neglecting this issue for a moment, averaged log-likelihood and its (approximated, equality with no missing data) gradient is:
$$F(\textbf{a})=\frac{1}{n} \sum_{k} \ln\left(\sum_{f\in B} a_f f(\textbf{x}^k) \right)$$
\be f\in B_C:\ \frac{\partial F(\textbf{a})}{\partial a_f}\approx \frac{1}{|K_C|} \sum_{k\in K_C}  \frac{f(\textbf{x}^k)}{\sum_{f\in B} a_f f(\textbf{x}^k)}\label{grad}\ee
The approximation is to still optimally use incomplete data sample here: average over all points having $C$ coordinates to improve modelling of correlations $C$.

The density in denominator can happen to be zero or negative, what should be prevented e.g. by initial reduction of coefficients. Likelihood optimization should lead to density positive in all points of the sample.
\subsection{The issue of negative densities}
Modelling density as linear combination $\rho=\sum_f a_f f$ has advantage of independent and inexpensive direct calculation of coefficients, however, being optimal for $L^2$ or likelihood it can still lead to negative densities, problematic in some applications. It can be avoided by using more complex parametrization, e.g. $\rho=(\sum_i a_f f)^p$ for some even $p$, or $\rho=\exp(-\sum_i a_f f)$, however, estimation of such parameters becomes much more complicated and costly, cannot be calculated independently.

Let us discuss importance and handling of the issue with eventual negative densities of $\rho=\sum_f a_f f$ parametrization:
\begin{itemize}
  \item Negative densities are relatively rare artifacts: require exceeding initial $\rho_\emptyset =1$. It is more likely for high order polynomial basis, which obtain large values at boundaries of the range - negative densities mainly happen near edges of $[0,1]^d$.
  \item Generally we can just use e.g. $\rho_\epsilon\equiv \max(\rho,\epsilon)$ for some $\epsilon>0$, repairing the problem at cost of approximated normalization.
  \item In many applications such rare negative density has nearly no meaning, especially if focused on more likely behavior. For example imputation using global maximum of density is always positive, expected value has to be in $[0,1]$. In such situations, negative density can only lead to a small error of predicted value. Searching for high probability path between two chosen points, such path should avoid negative density regions.
  \item Basic situation where negative density can be problematic is evaluation of probability of a given set of points, especially to search for outliers (single low probability points) or anomaly (change of statistics). Defining outliers as having density below some positive threshold, analogously to using $\max(\rho,\epsilon)$, negative density regions will be automatically also classified this way.
      Knowing the points of interest, we can make eventual repairment as part of query of density in a given point: if the asked point has turned out to have negative density, repair the modelled density before answering.
  \item Repairment of density basing on a point with negative density can be realized by example rescaling all coefficients (beside $\emptyset$), or reducing single coefficient with the largest value for this point. Alternative way is using gradient (\ref{grad}) of log-likelihood. Finding negative values of high dimensional polynomial is generally costly, edges of $[0,1]^d$ are natural candidates to test.
  \item In applications where negative values of density can be a crucial problem, reduction of probability of such situations can be obtained e.g. by additional likelihood maximization - ensuring positive density at least at the sample. Additionally, we can add e.g. $\xi\sum_f (a_f)^2$ to minimized function, getting regularization with lower coefficients and so smaller divergence from initial $\rho_\emptyset=1$.
\end{itemize}

\subsection{Normalization of variables to uniform $[0,1]$}
Due to normalization difficulty, the discussed approach requires working on finite ranges, conveniently $[0,1]^d$. We need to transform all variables to $[0,1]$, e.g. by simple rescaling for those which can be reliably bounded, or $\mathbb{R}\leftrightarrow [0,1]$ transformation like $s(y)=1/(1+e^{-y})$, $s^{-1}(x)=\ln(x/(1-x))$ if they cannot be bounded.

However, if variable after such transformation has very nonuniform distribution, fitting low order polynomials for decomposition of correlations might be ineffective, e.g. $j=1$ distinguish left and right part of $[0,1]$.

To prevent that, we can use estimated cumulative distribution function: $CDF:\mathbb{R}\to [0,1]$, $CDF(y)=\textrm{Pr}(Y\leq y)$ as this transformation. This way each separate variable $x=CDF(y)$ would have uniform distribution in $[0,1]$, e.g. 1/2 corresponds to median value. The $|C|=1$ terms should vanish this way - can be omitted, we can use only $|C|\geq 2$ terms: describing real correlations.

A basic approach to estimate $CDF$ for separate variables is to sort all its appearances $(l=|K_{\{i\}}|)$, then use $(y^k,k/l)$ as points for $CDF$: after some smothering, interpolation or spline, to be used (e.g. tabled) for transformation of original coordinates to $[0,1]$. Analogously $(k/l,y^k)$ points can be used to estimate $CDF^{-1}$ to transform back to the original variables.

\section{Comparison with cascade correlation}
Let us now discuss comparison of the presented approach with classical technique for modelling correlations in data sample: cascade correlations~\cite{cc}. It uses a fixed division of coordinates into input and output, we want to predict output from input, what can be imagined as imputation. For this purpose, it successively adds new neurons one-by-one, each one chosen to reduce the current prediction error (residue), then its weights are frozen - cannot be changed later due to complex dependencies.

Hierarchical correlation reconstruction can be seen as a single neuron corresponding to entire network in cascade correlation - above adding new neuron corresponds to adding new type of correlation to consider: function $f$ to basis $B$. Some its advantages:
\begin{itemize}
  \item Instead of heuristic greedy algorithm, we have real $L^2$ optimization with potentially complete basis - approaching the real joint probability distribution of the sample, with analytically controlled uncertainty, can be complemented for maximizing likelihood.
  \item Instead of dependent neuron weights which have to be frozen, coefficients of $\rho=\sum_f a_f f$ are independent: we can adapt all of them through the entire process, also freely add/remove functions from basis (considered correlation types), use only more certain coefficients for prediction while modelling on a larger basis, etc.
  \item Correlations allow to conclude in any direction - we can freely modify division into inputs/outputs, some original inputs might be missing.
  \item Coefficients have clear interpretation, e.g. $f_{11}$ determines increase/decrease of one variable with growth of the second, $f_{12}$ focus or spread of one variable with growth of the second, $f_{1j}$ analogously for approximately $j$-th cumulant. Combined with independence, it allows to directly use found coefficients for some further machine learning. Coefficients of $f_j$ polynomials have similar meaning as correspondingly: expectation value, variance, skewness, kurtosis, but can be directly translate to density.
  \item We directly estimate joint probability distribution, what additionally allows for example to estimate uncertainty of imputation, handle ambiguous imputation, or can be used for other applications requiring density, like clustering or morphing.
\end{itemize}
\section{Some applications}
As discussed, a basic example of application for such modeled density $\rho$ is imputation of missing coordinates, e.g. for some other machine learning algorithms. This approach additionally allows to handle problematic situation with ambiguous imputation - can provide uncertainty, a few most probable candidates, or entire probability distribution - for separate coordinates, or even joint probability distribution for all missing coordinates.

Such modelled density (e.g. as  polynomial) can be also directly used for various machine learning applications, especially that we can easily differentiate or integrate it, for example for multidimensional CDF as polynomial, generating random values from estimated density. Local density maxima can be interpreted as clusters in unsupervised learning. For supervised learning we can feed label to output coordinates then try to predict them (imputation), or we can model density for separate classes and classify comparing densities (Bayes). We can also directly use such density, e.g. for morphing: continuous transition between two points using high probability path, or to evaluate probability of a data point e.g. to detect outliers, anomaly (change of statistics), where we can tolerate missing some coordinates. We can also evaluate distance between found densities, e.g. using Kulback-Leibler divergence, or straightforward using some distance between their vectors of $a_f$ coefficients.

The found $a_f$ coefficients are independent and have clear meaning similar to cumulants, what allows to directly use them for some further machine learning. For example we could model correlations in a moving window - e.g. 1D for sequence or 2D for image. Estimating coefficients by using such window in all positions (like convolution), we can use their list as a fingerprint describing local correlations of given object, e.g. characteristic for the used sensor in image forensics.

Probability distribution is directly required in data compression, where already processed symbols should allow to model distribution of the current one. While standard approach is to use probability distribution tabled for various contexts, here we can try to directly reconstruct essential correlations with the context.

\begin{figure}[t!]
    \centering
        \includegraphics{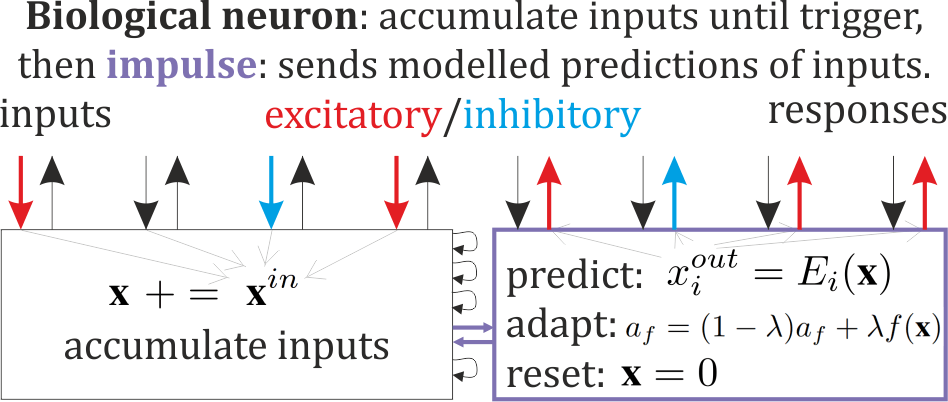}
        \caption{Application as a simple model of artificial neuron with analogous behavior as biological one: learning correlations and predicting signals, directly performing  e.g. shallow supervised learning. Each connection ($i$-th) has input and output (each can be connected to multiple neurons), neuron should predict output as its expected input: basing on remaining received inputs. Most of the time neuron accumulates excitatory ($x>1/2$) and inhibitory ($x<1/2$) inputs until some trigger condition, e.g. after a fixed time period, or gathering a number of signals, or some total strength (e.g. $|x-1/2|$) etc. Then for a moment it switches to impulse mode, when it uses the accumulated signals (normalized to $[0,1]$), and finally resets them. For each connection it predicts and produces input as it would be unknown for this connection: predicting from the remaining inputs ($E_i$ is modelled expected value (\ref{exp}) based on remaining coordinates). It also adapts the model (\ref{adapt}): coefficients of functions used in $\rho=\sum_{f\in B} a_f f$, maybe also adding new functions to the basis $B\ni f$, especially if creating a new connection. Missing data case allows for operation without signals from some inputs (still can predict outputs for all), and flexibility for adding new connections.}
       \label{neuron}
\end{figure}

Finally, flexibility in modelling and predicting inputs suggests to try to use the presented modelling as analogue for biological neuron, e.g. as discussed in Fig. \ref{neuron}. Building biology-inspired networks from such neurons could allow to transfer their learning and reasoning mechanisms. For such applications the number of connections might be large, what can be handled using a very sparse basis, adapted to the observed signal patterns - use coefficients only for characteristic patterns of correlations between $C$ connections: predicting any $i\in C$ connection from the remaining $|C|-1$. Such neuron maximizes local prediction capacities - based on local signals only, directly performing e.g. shallow supervised learning. The big question is building deep learning networks with them - how to choose architecture to learn long correlation chains with such neurons?

\bibliographystyle{IEEEtran}
\bibliography{miscites}
\end{document}